# An FPGA Implementation of a Time Delay Reservoir Using Stochastic Logic

LISA LOOMIS, Air Force Research Laboratory
NATHAN MCDONALD, Air Force Research Laboratory
CORY MERKEL, Air Force Research Laboratory

This paper presents and demonstrates a stochastic logic time delay reservoir design in FPGA hardware. The reservoir network approach is analyzed using a number of metrics, such as kernel quality, generalization rank, performance on simple benchmarks, and is also compared to a deterministic design. A novel re-seeding method is introduced to reduce the adverse effects of stochastic noise, which may also be implemented in other stochastic logic reservoir computing designs, such as echo state networks. Benchmark results indicate that the proposed design performs well on noise-tolerant classification problems, but more work needs to be done to improve the stochastic logic time delay reservoir's robustness for regression problems. In addition, we show that the stochastic design can significantly reduce area cost if the conversion between binary and stochastic representations implemented efficiently.

COMPUTING CLASSIFICATION SYSTEMS TERMS
Emerging architectures, Neural networks, Reconfigurable logic and FPGAs, Stochastic processes

KEYWORDS
Stochastic computing, reservoir computing, time delay reservoir, Bernstein polynomials

## 1 INTRODUCTION

Reservoir computing (RC) is proving to be a powerful machine learning technique for regression, classification, and forecasting of time series data. Introduced in the early 2000s by Jaeger [1] and Maass [2], RC is a type of neural network with an untrained recurrent hidden layer called a *reservoir*. A major computational advantage of RC is that the output of the network can be trained on the reservoir states using simple regression techniques, without the need for backpropagation. In the last decade and a half, RC been successful in a number of wide-ranging applications domains such as image classification [3], biosignal processing [4], and optimal control [5]. In some domains, RC has outperformed state-of-the-art techniques and is often easier to implement than methods such as Kalman filtering or long short term memory. Beyond its computational advantages, one of the main attractions of RC is that it can be implemented efficiently in hardware with low area and power overheads.

There are three major categories of RC. The first is echo state networks (ESNs) [1], where reservoirs are implemented using a recurrent network of continuous (e.g. logistic sigmoid) neurons. The second category, referred to as liquid state machines (LSMs) [2], utilizes recurrent connections of spiking (e.g. leaky integrate and fire) neurons. A challenge in both of these categories is routing. A reservoir with $N$ neurons will have up to $N^2$ connections, potentially creating a large area and power overhead. A third category of RC called time delay reservoirs (TDR) [6] avoids this overhead by time multiplexing resources. TDRs utilize a single neuron and a delayed feedback to create reservoirs with either a chain topology or even full connectivity (see Supplemental Material of [6]).

Besides a reduction in routing overhead, TDRs have two key advantages over ESNs and LSMs. First, additional neurons are added to the reservoir merely by increasing the delay in the feedback loop. Second, TDRs can use any dynamical system

This material is based upon work supported by the Air Force Office of Scientific Research (AFOSR) under award number FA9550-15RICOR122. Any opinions, findings and conclusions or recommendations expressed in this material are those of the author and do not necessarily reflect the views of AFRL. The material and results presented in this paper have been cleared for public release, unlimited distribution (Case Number 88ABW-2018-3440).
Author's addresses: Air Force Research Laboratory/ Information Directorate Rome, NY, USA.
Contains previously published material from C. Merkel, "Design of a time delay reservoir using stochastic logic: A feasibility study," *2017 International Joint Conference on Neural Networks (IJCNN)*, Anchorage, AK, 2017, pp. 2186-2192.

to implement their activation function and can easily be modeled via delay differential equations. This second point is particularly useful since it means that TDRs can be implemented using a variety of technologies. For example, in [6], Appeltant et al. used a Mackey-Glass oscillator, which models a number of physiological processes (e.g. blood circulation), as the non-linear node in a TDR. In [7], a TDR is demonstrated using a coherently driven passive optical cavity. A TDR has also been implemented using a single XOR gate with delayed feedback [8]. A common thread among all of these implementations is that they are analog and some, such as the photonic implementation, are still large prototypes that have yet to be integrated into a CMOS chip. Aside from the higher cost and design effort for analog implementations, they are much more susceptible to noise, which may degrade RC networks' performance, which are typically operated at the edge of chaos.

Digital RC designs, and digital circuits in general, have much better noise immunity compared to analog implementations. There have been a number of digital designs proposed for ESNs and LSMs, such as [9], but digital TDR designs are presently scarce. One example is given in [10], where the authors have implemented a Mackey-Glass-type TDR on an FPGA. One of the challenges with digital implementations is that the area cost can be high due to the requirement of multipliers for input weighting and implementation of the activation function. This is especially true if high precision is required. However, not all applications require high precision. An alternative design approach to conventional digital logic is stochastic logic, where values are represented as stochastic bit streams and characterized by probabilities. Stochastic logic has previously been used to implement ESNs [11, 12]. In this work, we explore the feasibility of implementing TDRs with stochastic logic. Specific contributions of this work are:

- A novel TDR design based on stochastic logic
- An FPGA implementation of the stochastic logic TDR design
- Analysis of the stochastic TDR design on two classification and two regression benchmarks
- Comparison of the stochastic TDR design to a deterministic design and an ESN

The rest of the paper is organized as follows. Section 2 discusses in detail time delay reservoir computing and stochastic logic. Section 3 presents a digital circuit implementation of a TDR based on stochastic logic. Section 4 presents our experimental results, both in simulation and FPGA hardware, followed by discussion of these results. Final conclusions are in Section 5.

## 2 BACKGROUND

### 2.1 Time Delay Reservoirs (TDR)

RC makes use of a random recurrent neural network to process time series data. The basic structure of an RC is shown in Figure 1. This is different from a traditional recurrent neural network (RNN) in that the hidden layer weights are randomly set and not trained, avoiding calculation-intensive techniques such as backpropagation. Time series data in the *input layer* are multiplied by a random, fixed weight vector, added to a random bias vector, and then used as inputs to the *reservoir layer*. The reservoir layer is a recurrent neural network composed of $N$ neurons that are connected by a second random, fixed weight matrix. In the case of an ESN, the neurons' activation functions are logistic sigmoid. In the case of an LSM, the neurons have spiking outputs. The state of the reservoir layer (the outputs of all of the neurons) is connected to an *output layer* via a third weight matrix which is trained to match target outputs via linear regression. The weights in the input and reservoir layers serve to randomly project the inputs into a high-dimensional space which increases their linear separability. In addition, the recurrent connections of the reservoir layer provide a short-term memory that allows inputs to be integrated over time. This is critical to analyzing data based on its behavior over multiple timesteps.

A TDR is a special type of RC that shares resources in time to reduce routing overhead. The primary difference between a conventional RC (Figure 1) and a TDR (Figure 2) is that TDRs only use a single neuron in the reservoir layer. The single neuron is shared in time by sampling and holding each input for $N$ timesteps. At each timestep, one element of the reservoir's input is computed by multiplying the sampled data by one of the input weights and adding one of the bias terms. Then, this result is added to a delayed version of the neuron's output, which has been delayed by $\tau$ timesteps. The final summation is fed back into the neuron, and the process starts over for the next timestep. After $N$ timesteps, a new input is sampled. In this work, we use $\tau = N + 1$. In this case, it can be shown that the TDR is equivalent to a conventional RC with a ring topology in the reservoir layer. The next section will present an efficient hardware implementation of a TDR using stochastic computing techniques.

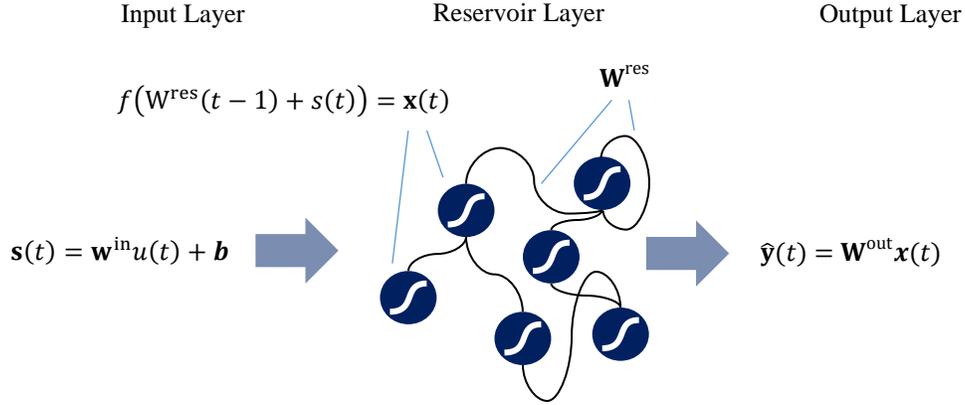

Fig. 1. Basic structure of an RC design, showing the three layers: input, reservoir, and output. $u$ is a time series input; $\mathbf{w}^{in}$ is a random, fixed weight vector; $\mathbf{W}^{res}$ is a random, fixed weight matrix; $\mathbf{b}$ is a random, fixed bias; $\mathbf{s}$ is the input to the reservoir layer; $\mathbf{x}$ is a vector of reservoir states; $f$ is the neuron activation function; $\mathbf{W}^{out}$ is the trained output weight matrix; and $\hat{\mathbf{y}}$ is the output.

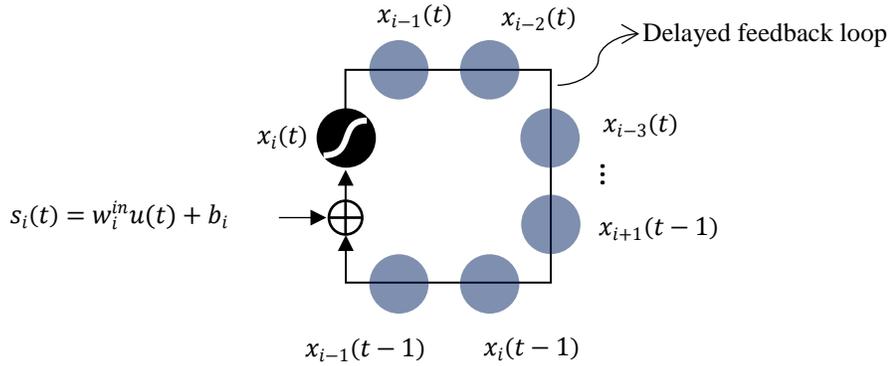

Fig. 2. Overview of a TDR, showing the use of a delay line to create multiple virtual reservoir nodes from one physical node.

2.2 Stochastic Logic

Stochastic logic was pioneered by von Neumann [13] half a century ago and was later adopted by the machine learning community to improve hardware complexity, power, and reliability [14, 15]. At the heart of stochastic logic is the stochastic representation of real numbers. In a single-line bipolar [15] stochastic representation, a number $q$ between -1 and 1 is represented as the probability $p$ of a length-$L$ Bernoulli process, or bit stream $z$ [16, 17]. Probabilities below 0.5 represent negative numbers while probabilities above 0.5 represent positive numbers. Furthermore, probability values of 0, 0.5, and 1 represent the values -1, 0, and 1. Therefore, we have the mapping $p = (q+1)/2$. For example, the number $q = -0.5$ maps to $p = 0.25$, which can be represented by any bit stream where the expected number of 1's is 25%, e.g. $z = 00101000$. Bit streams can be generated by comparing the original number with random numbers between -1 and 1 using a linear feedback shift register (LFSR). If the random number is less than or equal to $q$, then a 1 is placed in the bit stream, otherwise a 0. As $L$ becomes large, the representation becomes more accurate by reducing sampling noise and increasing precision. Converting from a stochastic representation back to a number can be achieved by counting the number of 1's and 0's in the bit stream. By initializing a counter to 0, adding 1 every time a '1' is encountered in the stream and subtracting 1 every time a '0' is encountered, the final counter value will be the two's complement binary representation of the bit stream.

One advantage of a stochastic representation is that several mathematical operations become trivial in hardware. Implementations of several stochastic logic operations are shown in Figure 3. Multiplication and negation can be implemented with XNOR gates and inverters, and multiplexers can be used to implement weighted summation. Furthermore, it is possible to implement Bernstein polynomials by connecting an adder to the multiplexer select line [20]. The inputs to the adder are $n$ copies of the polynomial argument, each with a statistically independent stochastic representation. Here, $n$ is the order of the polynomial. The multiplexer's data inputs are the stochastic representations of the Bernstein coefficients. In this work, we use Bernstein polynomials to approximate the reservoir neuron's activation function.

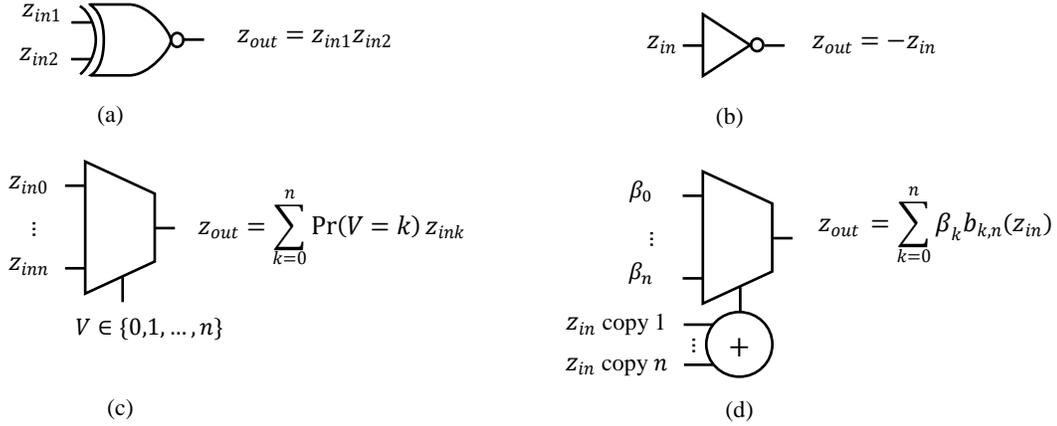

Fig. 3. Basic logic gates implementing mathematical operations on single-line bipolar stochastic bit streams: (a) multiplication, (b) negation, (c) weighted summation, (d) Bernstein polynomial, where $\beta_k$ are the Bernstein coefficients and $b_{k,n}$ are the $n$th order Bernstein basis polynomials.

## 3 STOCHASTIC LOGIC TDR CIRCUIT DESIGN

The stochastic logic TDR implemented in this work is shown in Figure 4. The design is composed of three parts to provide input weight, compute the non-linear activation function, and hold the reservoir state. The input weighting stage takes in an analog signal, converts it to a digital signal using an analog-to-digital (A2D) converter, and converts that to a stochastic bit stream using a binary-to-stochastic (B2S) converter. In this design, the number of bits in the LFSR in each B2S is equal to the number of bits in the binary representation of the input $q$. Each timestep, the B2S compares $q$ to the LFSR's current value and produces one bit of the stochastic representation of $q$. The stochastic representation of the input is then multiplied by the input weight using an XNOR gate, as discussed in the Section 2.2. Then, the signal is mixed with the delayed reservoir state and added to the input bias using MUXes. The non-linear node estimates the non-linear activation function $f(s)$ using Bernstein polynomials. Shift registers are used to delay the non-linear node's input in order to create multiple statistically independent copies of the same stochastic bit stream. In this design, the activation function implemented is $f(s) = \sin^2(\gamma s)$, where $\gamma$ is a frequency term. However, because Bernstein polynomials map the unit interval to the unit interval, we shift and scale $f(s)$ such that it lies entirely in the unit square. Figure 5 shows $f(s)$ and the stochastic logic approximation using Bernstein polynomials with $L = 1000$ and $n = 10$. Notice that the features of the curve around $s = 0.5$ are not reproduced well by this approximation but could be by increasing $n$. However, it was found that the approximation works well for some of the benchmarks explored in this work.

After the activation function is computed, the reservoir node $x_i$ is converted back to a binary number using a stochastic-to-binary (S2B) converter and then placed in a shift register which holds the entire reservoir state $x$. Although the states could be stored in their stochastic representation, storing them as binary values is more area-efficient since $L \gg q$. Note that a control block is also included in Figure 4 to emphasize that the sample-and-hold circuit (inside the A2D), the B2S, and the S2B require a state machine when implemented in hardware. The shift register serves as the delay line shown in Figure 2. Training was performed using linear least-squares regression.

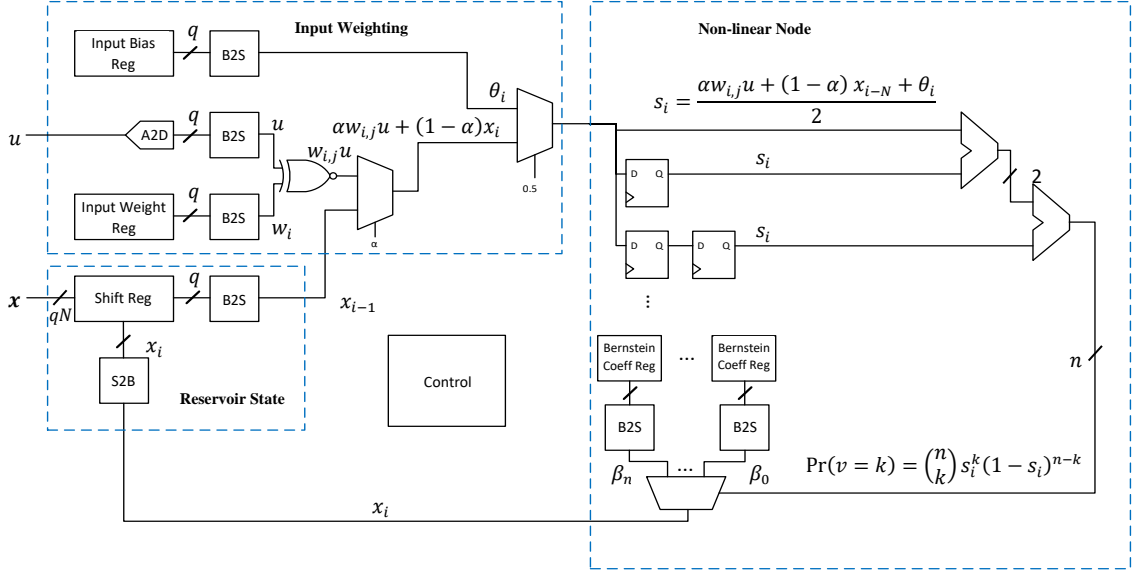

Fig. 4. Proposed stochastic logic TDR.

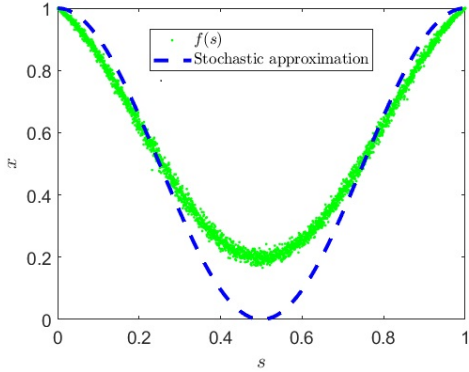

Fig. 5. Sine squared activation function and stochastic approximation implemented using Bernstein polynomials ($n = 5$, $L = 1000$).

Following the choice of $N$ and $L$, the reservoir attenuation parameter $\alpha$ and the activation function non-linearity parameter $\gamma$ are the principal design parameters for the stochastic logic TDR. One way to select appropriate values is to consider application-dependent metrics such as accuracy, specificity, sensitivity, mean squared error, etc., and see how they vary over the parameter space via, e.g. a grid search. Another way is to use metrics that capture features such as the reservoir's short-term memory capacity, ability to linearly separate input data, and capability of mapping similar inputs to similar reservoir states (generalization) while mapping different inputs to distant reservoir states (separation). These types of application-independent metrics provide more insight into the effects of different parameter choices on the TDR's computing power than metrics like accuracy. This work makes use of such metrics: kernel quality (KQ) and generalization rank (GR) [21]. KQ is calculated by driving the reservoir with $N$ random input streams of length $m$. At the end of each sequence, the reservoir's final state vector is inserted as a column into an $N \times N$ state matrix **X**. Then, KQ is equal to the rank of **X**. It is desired to have **X** be full rank, meaning that different inputs map to different reservoir states. However, note that the number of training patterns is usually much larger than $N$, so if rank(**X**) = $N$, it doesn't mean that any training dataset can be fit exactly. In fact, exact fitting, or *overfitting*, is generally bad, since it means that the TDR (or any machine learning algorithm) will not generalize well to novel inputs. Therefore, another metric, GR, is used to measure the TDR's generalization capability. GR is calculated in a similar way, except that all of the input vectors are identical except for some random noise. GR is an estimate of the Vapnik-Chervonenkis dimension [21], which is a measure of learning capacity. A low GR is desirable. Therefore, to achieve good performance, the difference KQ-GR should be maximized.

Figure 6 shows the KQ and GR metrics for the stochastic logic TDR with $L = n = \infty$ (i.e. a deterministic TDR). The values are normalized to $N$ and are averaged over 10 runs. In each subplot, the size of the reservoir is $N = 50$ and $m = 50$. KQ (Figure 6(a)) is close to 0 for $\alpha = 0$. When $\alpha = 0$, the TDR does not accept any new inputs, so if the initial TDR state is all zeros, then the final state matrix will be a zero matrix, which has a rank of zero. For larger $\alpha$ values, KQ becomes non-zero. When $\gamma$ is small, the activation function is approximately linear, which leads to a smaller KQ. In fact, it is likely that the TDR operates in the deterministic phase for $\gamma < 1$. As $\gamma$ becomes larger, KQ becomes equal to $N$. This is because the non-linearity of the activation function increases with $\gamma$ and results in the TDR operating within the chaotic regime. The GR metric (Figure 6(b)) has similar behavior, though recall that a low GR value is desirable. When the difference KQ-GR is taken (Figure 6(c)), a small region of optimal $\alpha$ and $\gamma$ is observed. If $\alpha$ is too large, then the TDR will have no memory; and if $\alpha$ is too small, then it will ignore inputs. Furthermore, if $\gamma$ is too large, then the TDR will overfit the training data; and if $\gamma$ is too small then the TDR will not have enough dynamic behavior. In practice, the optimal parameter values will have some application dependence. The results in Figure 6(c) provided us with a starting point for determining good values of $\alpha$ and $\gamma$. Further analysis determined $\alpha = 0.6$ and $\gamma = 2$ to be the best for the studied benchmarks. However, from Figure 6(c), this parameter choice was expected to be suboptimal. Therefore, while one can generally consider such metrics such as GR and KQ, one ought to optimize the behavior of the RC for a chosen set of applications.

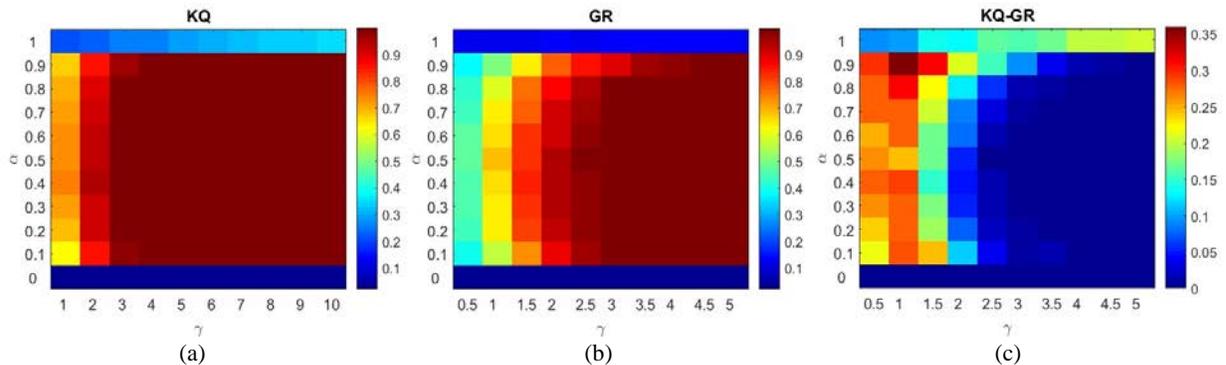

(a)  (b)  (c)

Fig. 6. Computational capability of TDR for $L = \infty$, $N = 50$. (a) KQ vs. $\alpha$ and $\gamma$. (b) GR vs. $\alpha$ and $\gamma$. (c) KQ-GR vs. $\alpha$ and $\gamma$.

Next we studied the effect of the stochastic bit stream length $L$ on the TDR metrics. Intuitively, one would expect that a small value of $L$ would lead to both a large KQ and GR, since the variance introduced from the stochastic computation is proportional to $1/L$. This was indeed observed in the KQ and GR metrics for two cases (Figure 7).

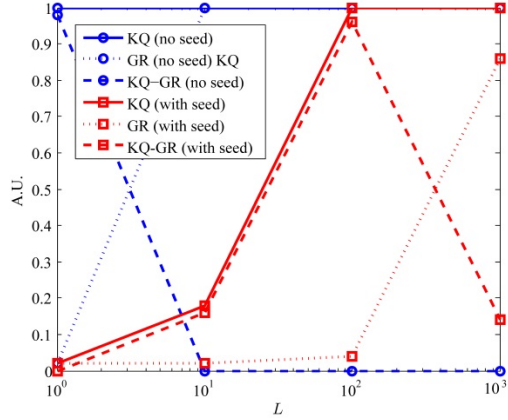

Fig. 7. KQ and GR vs. *L* in the stochastic logic TDR. Results are shown for the cases where the PRNG is not (*no seed*) and is (*with seed*) re-seeded for each reservoir node.

In the first case, each LFSR was only seeded at the beginning of the simulation. This resulted in KQ-GR equal to zero over all *L* values, except *L* = 1, where the noise was not large enough to modify the stochastic representation. From the previous discussion, we see that KQ-GR will eventually become non-zero as $L \rightarrow \infty$. However, that would mean that the TDR may have to wait an impractical number of clock cycles for each calculation. Instead, the approach used in this work was to re-seed each PRNG for every reservoir node with a unique seed for that node. Although this does not eliminate stochastic noise, it does keep the effect of the noise approximately constant for each node. With re-seeding, KQ-GR is non-zero for reasonable *L* values such as 100 and 1000.

Hardware experimentation in this work was performed on a Xilinx Spartan-6 XC6SLX75T board. Written in VHDL, the stochastic reservoir design followed the structural layout of Figure 4. A fixed-point deterministic TDR was also developed to run on the Xilinx FPGA for comparison. Eight-bit signed fixed-point numbers represent the range (-1, 1). This gives similar precision to L = 128 in the stochastic design (see Section 4). The design is broken into the same three blocks as the stochastic design: input node, nonlinear node, and reservoir state. The nonlinear node is implemented as a piecewise linear approximation of $\sin^2(\gamma s)$ function. The same parameters α = 0.6 and bias $\theta$ =0.6 are used. A randomly-generated mask is hard-coded into the input node, so there are no sources of randomness in the design.

## 4 RESULTS AND DISCUSSION

The stochastic logic TDR proposed in this work was tested on four standard benchmark tasks: NARMA10, sine/square wave discrimination, the Santa Fe laser dataset, and non-linear channel equalization. For each benchmark, we compared the stochastic TDR MATLAB simulation for $L$ = 16, 32, 64, and 128 (referred to as SS16, SS32, SS64, SS128 in the plots), the stochastic TDR on the FPGA (SF16, SF32, SF64, SF128), the deterministic TDR MATLAB simulation (DS), the deterministic TDR on the FPGA (DF), as well as the performance of an ESN. In addition, each test case was run for reservoir sizes of $N$ = 20, 30, 40, and 50. Here, we have omitted the cases of ($N = 40, L = 128$), ($N = 50, L = 64$), and ($N = 50, L = 128$) for the FPGA since they would not fit on the board. Training and testing were performed on 1000 points each, and results were taken over 20 trials, except for Santa Fe laser task, which has one training set of 9000 points and one testing set of 1000 points. Data points are sent to the board, and the reservoir states are sent back. Training of $\mathbf{W}^{\text{out}}$ is done in MATLAB. For *N* = 40 and *N* = 50, the circuit with larger values of *L* could not fit on the Xilinx board, because of the way the reservoir output is stored before being transmitted.

### 4.1 NARMA10

NARMA10 is a standard benchmark used in RC research. Given a random input sequence $u(t) \in [0,0.5]$, the goal is to train the RC to compute

$$y(t+1) = 0.3y(t) + 0.05y(t)\left[\sum_{k=0}^{9} y(t-k)\right] + 1.5u(t)u(t-0) + 0.1 \tag{13}$$

In this work, the TDR was trained on a set of 1000 points and tested on the subsequent 1000 points of the function. Figure 8 shows the normalized mean square error (NMSE) of the TDR for the test data, where a low value is desirable. The NMSE is calculated as

$$\text{NMSE} = \frac{\sum_t (y(t) - \hat{y}(t))^2}{\sum_t (y(t) - \langle y \rangle)^2}, \tag{14}$$

where $\langle \cdot \rangle$ is the arithmetic mean.

The stochastic and deterministic TDR simulations perform closest to the ESN for all values of $N$. The TDRs run on the board perform poorly. The NARMA10 task requires high precision for good performance, which the fixed-point implementation does not provide.

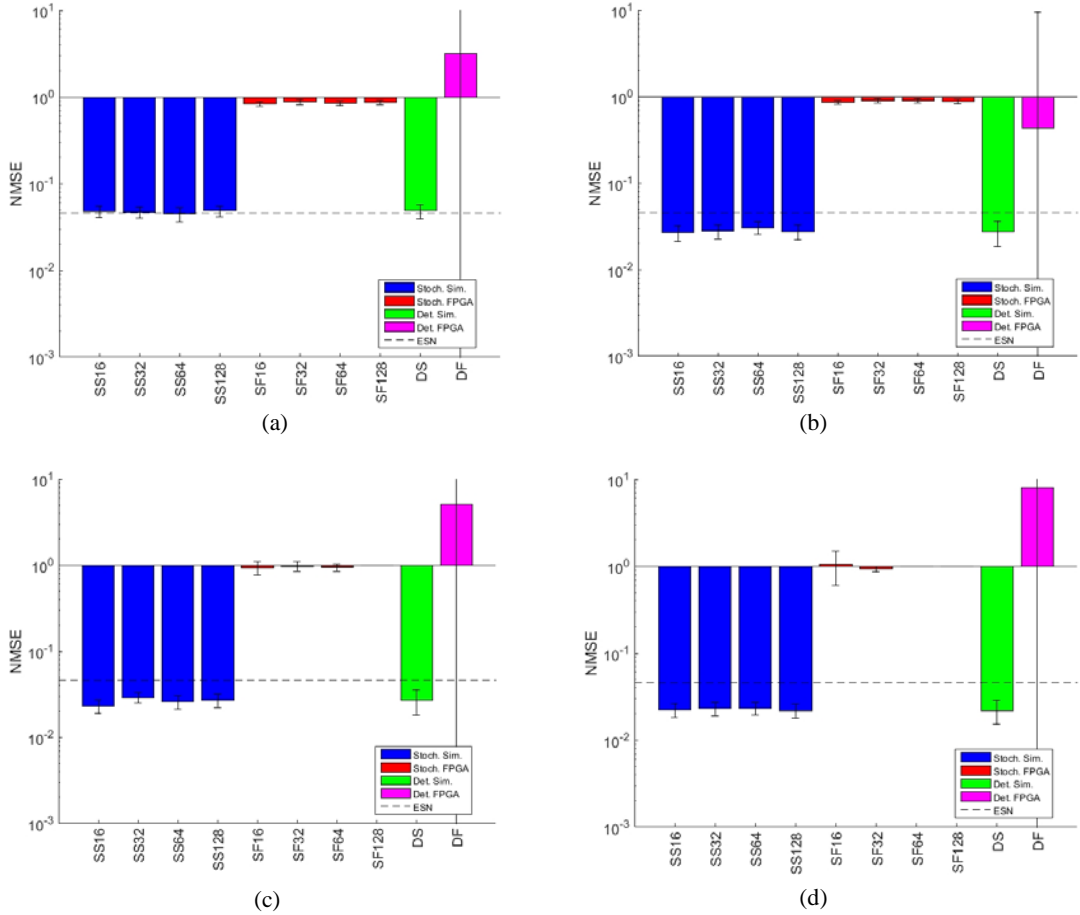

Fig. 8. NARMA10 prediction NMSE results for (a) $N = 20$, (b) $N = 30$, (c), $N = 40$, and (d) $N = 50$. Error bars represent the standard deviation

## 4.2 Sine vs. Square Wave Discrimination

To test its performance on a time-series binary classification task, the stochastic logic TDR was trained to discriminate between sine and square wave signals. 20 signals of random wave sequences of 1000 points were used for training and another 20 signals were used for testing, as was the case for the NARMA10 benchmark. Task performance is measured by the classification error. The stochastic TDR simulations perform closest to the ESN for all values of $N$, with nearly 100% classification accuracy. Note that in some cases the stochastic design significantly outperforms the deterministic TDR. This is likely a result of the stochastic noise preventing overfitting. We see little variation in the error with the bit stream length. However, if $L$ is reduced substantially (e.g. to just 2 or 4), then the error increases towards random chance (50%). In addition, there is a general decreasing trend in the error with increasing $N$, except in the case of the case of the stochastic TDR run on the FPGA.

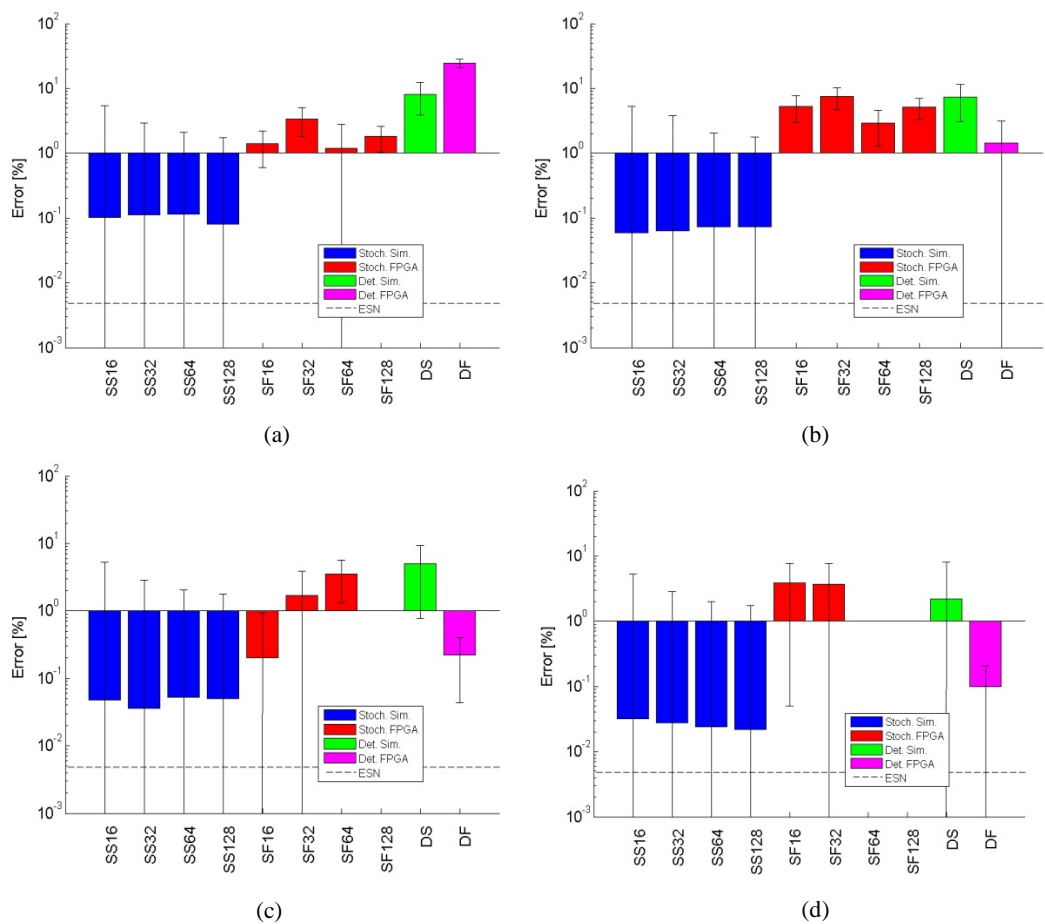

Fig. 9. Sine vs. square wave discrimination results for (a) $N = 20$, (b) $N = 30$, (c), $N = 40$, and (d) $N = 50$. Error bars represent the standard deviation.

4.3 Santa Fe Laser

Another regression benchmark is the Santa Fe laser dataset, where the network must predict a chaotic laser power sequence [22]. The network is trained on the first 9,000 points and then must predict the next 1,000 points, where a low NMSE is again desired. Here, none of the TDR implementations perform as well as the ESN. The parameters of the TDR were selected for their performance on the sine/square wave discrimination task, and may be poorly suited for this prediction task.

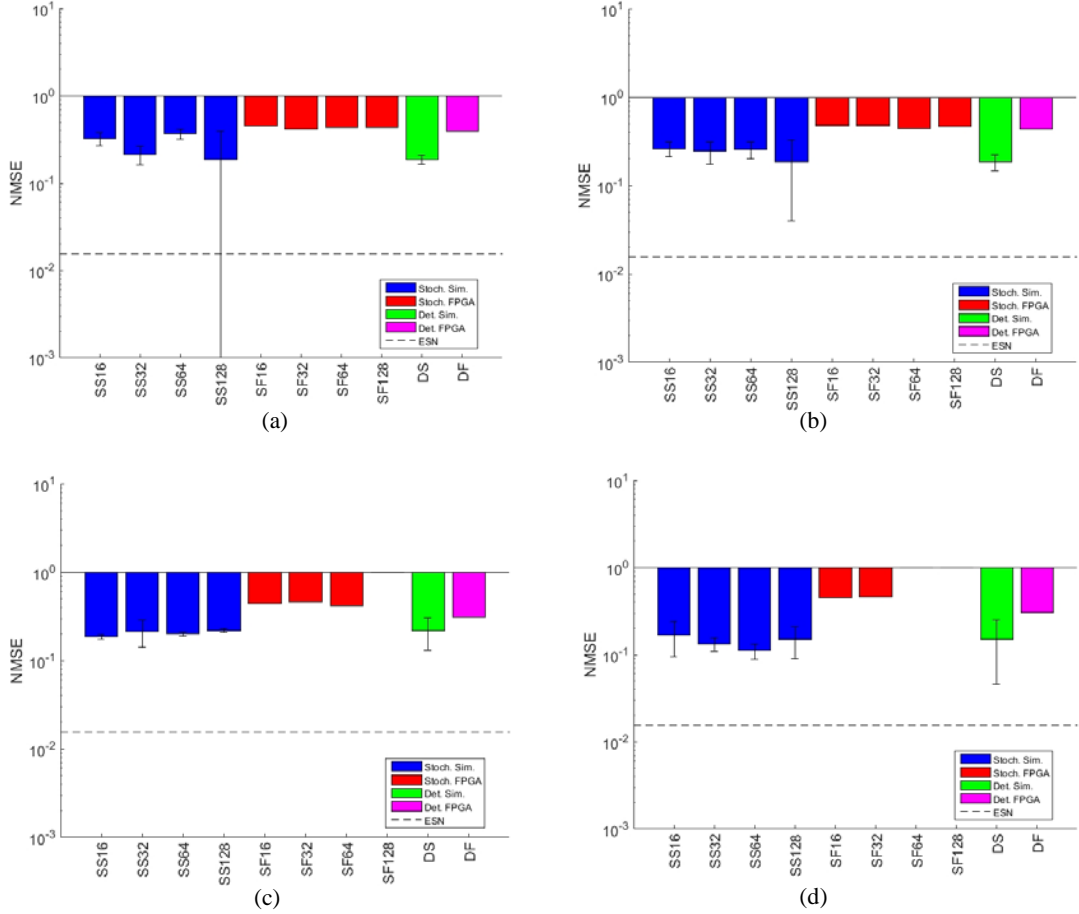

Fig. 10. Santa Fe Laser prediction NMSE results for (a) $N = 20$, (b) $N = 30$, (c), $N = 40$, and (d) $N = 50$. Error bars represent the standard deviation.

4.4 Non-Linear Channel Equalization

Lastly, the non-linear channel equalization task models a wireless communication channel input signal $d(n)$ traveling through multiple paths to a nonlinear and noisy receiver [23]. The task is to reconstruct the original input $d(n)$ from the output $u(n)$ of the receiver. Task performance is again measured using symbol error rate (SER), where SER < 1e-3 is possible for ESNs. Here, the simulated stochastic TDR performs closest to the ESN result. Performance improves slightly with increasing $N$.

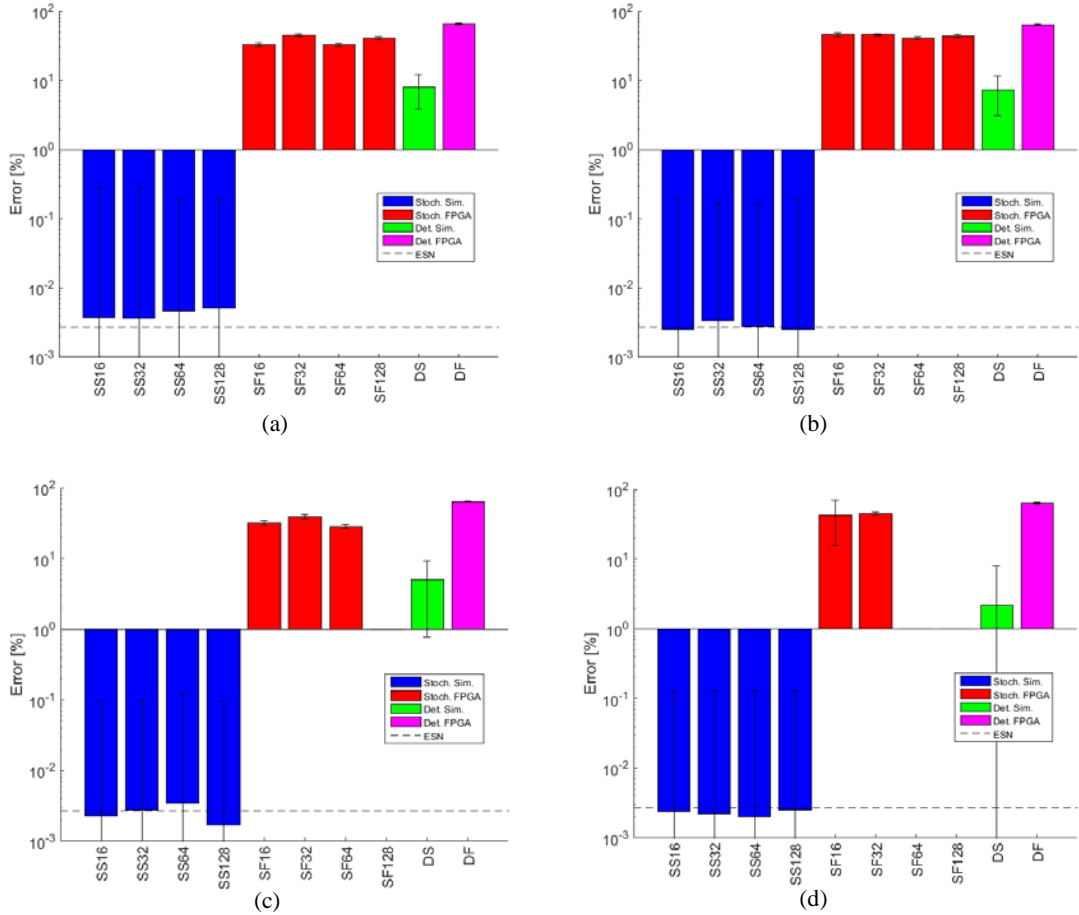

Fig. 11. NCE results for (a) $N = 20$, (b) $N = 30$, (c), $N = 40$, and (d) $N = 50$. Error bars represent the standard deviation.

Overall the stochastic TDR performs well in simulation on classification tasks such as sine/square classification and the NCE task. On tasks requiring precision, such as the NARMA10 and Santa Fe Laser predictions, NMSE is higher than the deterministic simulation. On these benchmarks, the stochastic TDR run on the board performs worse than the stochastic TDR simulated in MATLAB. The simulated stochastic reservoir and the one run on the FPGA are desired to be the same. However, some inconsistencies are due to inherent differences between coding in MATLAB and in VHDL, but future work will seek to eliminate them. In the two classification tasks, we also notice that the stochastic simulation usually outperforms the deterministic simulation. This is a nice byproduct of the sampling noise that comes from the stochastic representation, which acts as a regularizer, preventing overfitting to the training data.

Area and power estimates were performed with Xilinx's mapping report and XPower Analyzer. To communicate with the TDR on the Xilinx board requires additional registers devoted to communicating with a UART protocol and storing the outgoing data. The results in Figures 12 and 13 account for the TDR design without this storage. The Xilinx board has many gates that are not used in the reservoir design, making the quiescent power higher than the dynamic power in both the stochastic and deterministic implementations. A custom solution such as an ASIC would be a better choice for decreased power and area. We expected the quiescent power consumption to be much smaller for the stochastic design, since stochastic logic functions typically require fewer resources. However, as we show below, the conversion between binary and stochastic numbers has significant area overhead when implemented using LFSRs and counters. Since quiescent power is proportional to gate usage, the stochastic TDR's quiescent power became comparable to the deterministic design. Interestingly, we also noticed that the dynamic power of the stochastic logic TDR actually decreases with the bit stream length. This is counterintuitive, but can be explained as follows. Imagine that we have a stochastic signal with probability of 0.5. For a bit stream length of 2, there are two ways to represent it: 01, or 10. Both of these have one transition. The dynamic power is proportional to the number of transitions divided by $L$, which is 1/2 in this case. Now, if we let $L = 10$, we have several options to represent 0.5. The worst-case would be a bit stream like 0101010101, where there is a maximum number of transitions. Here, our dynamic power is proportional to 9/10. However, this is a relatively unlikely bit stream. A more likely scenario is a bit stream with just one or two transitions, such as 0000011111 or 0011111000, which have dynamic powers proportional to 1/10 and 2/10 respectively. In other words, there are more power-efficient (dynamic power) ways to represent a given probability when the bit stream is longer.

Figure 13 compares the number of FPGA slices that were utilized in our designs. Higher $N$ values require a longer delay register, leading to both longer calculation times and more occupied slices. The explored configurations took 1-2% of the available slices, when not accounting for the UART module and registers used to store the reservoir output values. The storage of output values takes up most of the slices on the FPGA board, limiting the size of the reservoir in this implementation.

The deterministic reservoir requires slightly more slice LUTs (look-up tables) than the stochastic implementation at all $N$, and many fewer slice registers. However, the LFSRs in the stochastic design add 16 registers apiece. A more area-efficient way of implementing pseudo-random numbers would make the stochastic design competitive with the deterministic area-wise. In addition, to add precision to the deterministic design would require a larger area, while an increase in $L$ in the stochastic reservoir would increase the calculation time but not the circuitry.

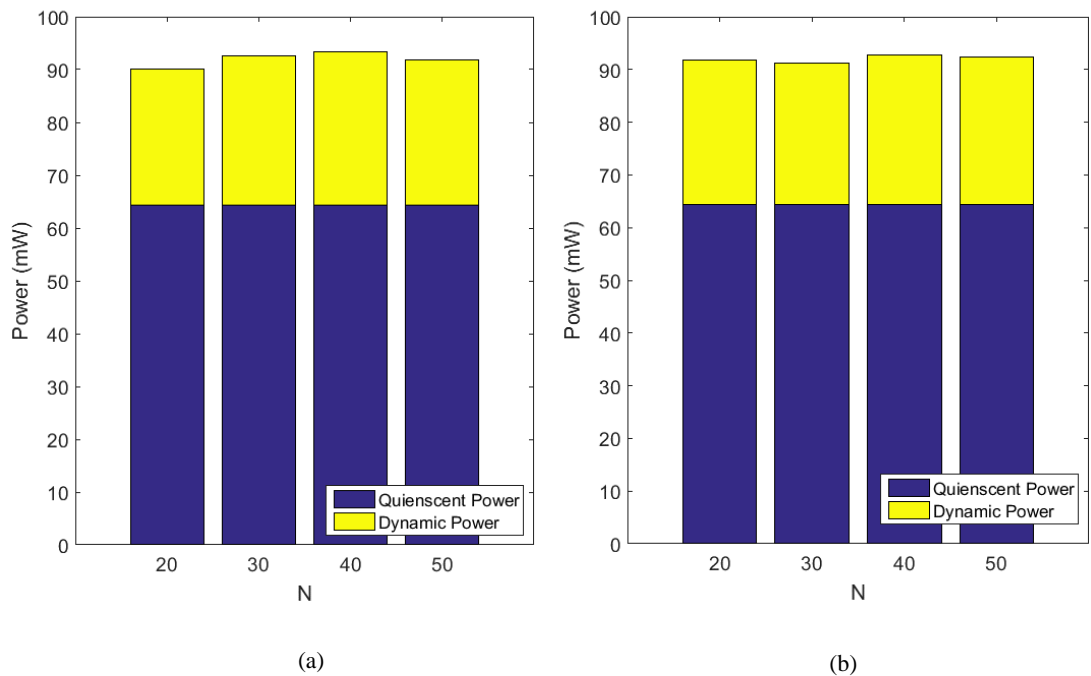

Fig. 12. Power estimation of (a) stochastic design and (b) fixed-point deterministic design.

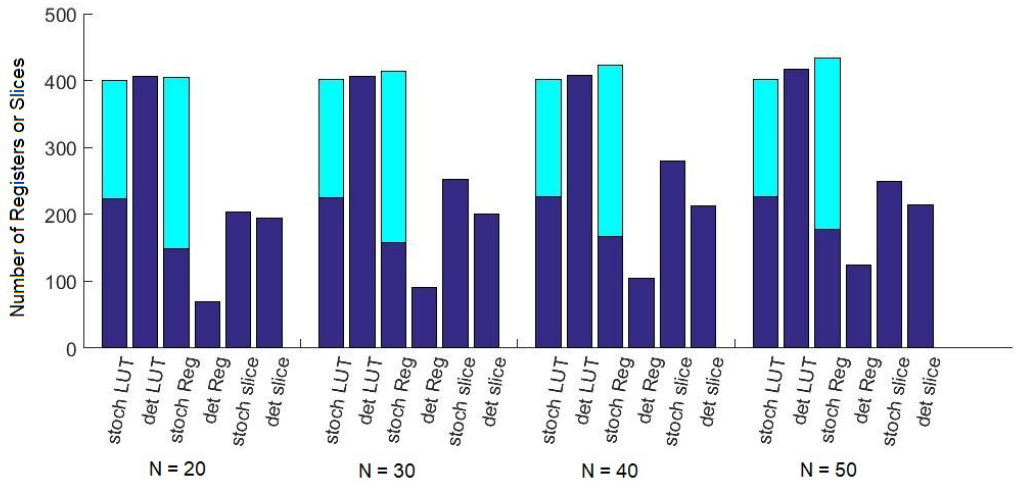

Fig. 13. Area usage of stochastic and fixed-point deterministic designs. Light blue segments on the stochastic designs show resources utilized for conversion between stochastic and binary representations.

## 5   CONCLUSIONS

This paper presents and demonstrates a stochastic logic time delay reservoir design implemented in FPGA hardware. The reservoir network approach is analyzed using a number of metrics, such as kernel quality, generalization rank, and performance on simple benchmarks. The use of a Bernstein polynomial as the non-linear node allows any activation function to be approximated, adding flexibility to the design. A novel re-seeding method is introduced to reduce the adverse effects of stochastic noise. Benchmark results indicate that the proposed design performs well on noise-tolerant classification problems, but more work needs to be done to improve the stochastic logic time delay reservoir's robustness for regression problems. Further research will be done to increase the resource efficiency of our FPGA hardware implementation and apply the TDR to more computationally complex tasks.

ACKNOWLEDGMENTS

This material is based upon work supported by the Air Force Office of Scientific Research (AFOSR) under award number FA9550-15RICOR122. Any opinions, findings and conclusions or recommendations expressed in this material are those of the author and do not necessarily reflect the views of AFRL. The material and results presented in this paper have been cleared for public release, unlimited distribution (Case Number 88ABW-2018-3440).